\pdfoutput=1

\documentclass[11pt]{article}

\usepackage{acl}

\usepackage{times}
\usepackage{latexsym}

\usepackage[T1]{fontenc}

\usepackage[utf8]{inputenc}

\usepackage{microtype}

\usepackage{inconsolata}

\usepackage{array}
\usepackage{enumitem}
\usepackage{wrapfig}
\usepackage{multirow}
\usepackage{graphicx}
\usepackage{amsfonts}
\usepackage{duckuments}
\usepackage{booktabs}
\usepackage{amsmath}
\usepackage{hyperref}
\usepackage{pdfpages}

\title{Language is All a Graph Needs}

\title{
\vspace*{-0.5in}
{{\small \hfill EACL 2024}\\
\vspace*{.25in}}
Language is All a Graph Needs}

\author{
Ruosong Ye\textsuperscript{\rm 1},
Caiqi Zhang\textsuperscript{\rm 2},
Runhui Wang\textsuperscript{\rm 1},
Shuyuan Xu\textsuperscript{\rm 1},
Yongfeng Zhang\textsuperscript{\rm 1} \\
\textsuperscript{\rm 1}Department of Computer Science, Rutgers University, New Brunswick, US\\
\textsuperscript{\rm 2}Language Technology Lab, University of Cambridge, UK\\
\texttt{ruosong.ye@rutgers.edu, cz391@cam.ac.uk, runhui.wang@rutgers.edu, }\\
\texttt{shuyuan.xu@rutgers.edu, yongfeng.zhang@rutgers.edu}
}

\begin{document}
\maketitle
\begin{abstract}
The emergence of large-scale pre-trained language models has revolutionized various AI research domains. Transformers-based Large Language Models (LLMs) have gradually replaced CNNs and RNNs to unify fields of computer vision and natural language processing. Compared with independent data samples such as images, videos or texts, graphs usually contain rich structural and relational information. Meanwhile, \textbf{language}, especially natural language, being one of the most expressive mediums, excels in describing complex structures. However, existing work on incorporating graph problems into the generative language modeling framework remains very limited. Considering the rising prominence of LLMs, it becomes essential to explore whether LLMs can also replace GNNs as the foundation model for graphs. In this paper, we propose \textbf{InstructGLM} (\textbf{Instruct}ion-finetuned \textbf{G}raph \textbf{L}anguage \textbf{M}odel) with highly scalable prompts based on natural language instructions. We use natural language to describe multi-scale geometric structure of the graph and then instruction finetune an LLM to perform graph tasks, which enables \textbf{Generative Graph Learning}. Our method surpasses all GNN baselines on ogbn-arxiv, Cora and PubMed datasets, underscoring its effectiveness and sheds light on generative LLMs as new foundation model for graph machine learning.  Our code is available at \href{https://github.com/agiresearch/InstructGLM}{https://github.com/agiresearch/InstructGLM}. 
\end{abstract}

\section{Introduction}
Prior to the advent of Transformers \cite{vaswani2017attention}, various artificial intelligence domains with different inductive biases had diverse foundational model architectures. For instance, CNNs \cite{lecun1995convolutional,szegedy2016rethinking} were designed with considerations for spatial invariance in images, leading to superior performance in computer vision tasks \cite{deng2009imagenet,lin2014microsoft}. Memory-enhanced models like RNNs \cite{elman1990finding} and LSTM \cite{hochreiter1997long,cho2014learning} were widely used for handling sequential data such as natural language \cite{sarlin2020superglue} and audio \cite{chen2021investigation}. Graph Neural Networks (GNNs) have long been the preferred choice in graph learning due to their proficiency in capturing topological information through message passing and aggregation mechanisms \cite{kipf2016semi,velivckovic2017graph,hamilton2017inductive,han2023alternately}.

In recent years, the AI community has witnessed the emergence of numerous powerful pre-trained Large Language Models (LLMs) \cite{devlin2018bert,raffel2020exploring,brown2020language,touvron2023llama,ouyang2022training}, which are driving huge advancements and lead to the pursuit of Artificial General Intelligence (AGI) \cite{ge2023openagi,bubeck2023sparks}. Under this background, there is a trend towards unification in model architectures across different domains. Specifically, pre-trained Transformers have demonstrated remarkable performance on various modalities, such as images \cite{dosovitskiy2020image} and videos \cite{arnab2021vivit} in computer vision, text in natural language processing \cite{singh2021com2sense}, structured data in graph machine learning \cite{ying2021transformers}, personalized data in recommender systems \cite{geng2022recommendation}, decision sequences in reinforcement learning \cite{di2023towards}, and visual-text pairs in multimodal tasks \cite{radford2021learning}. There has even been Transformers capable of handling twelve modalities \cite{zhang2023meta}.

\begin{figure*}
\vspace{-40pt}
	\centering
	\includegraphics[width=1.0\linewidth]{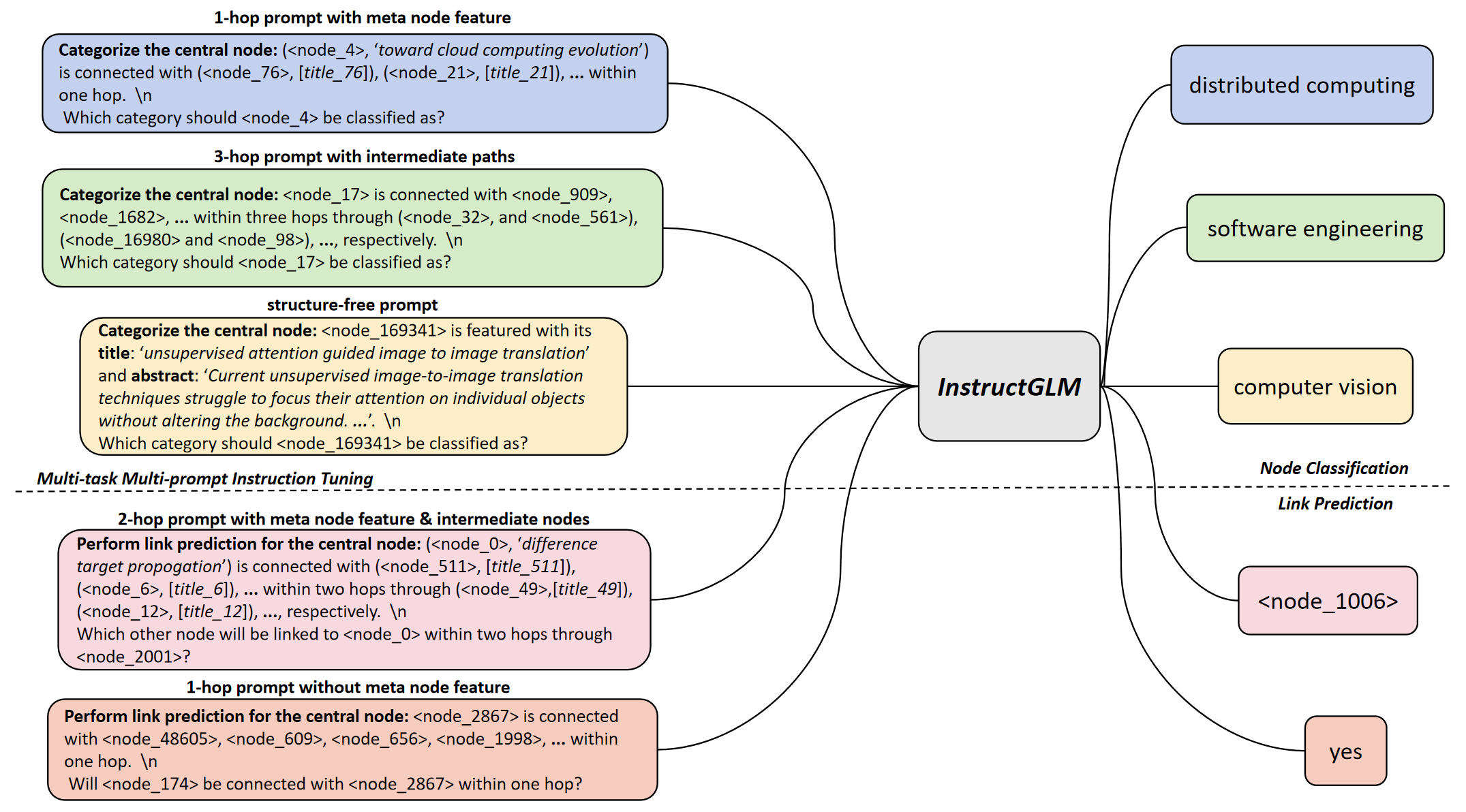}
	\caption{Illustration of the InstructGLM Framework. We fine-tune InstructGLM under a Multi-task Multi-prompt instruction tuning framework, enabling it to solve various graph machine learning tasks with the structure information purely described by natural language.}
	\label{fig_1}
 \vspace{-15pt}
\end{figure*}

Alongside advancements in model architectures, there is also a noteworthy trend towards the adoption of unified processing techniques for multimodal data. T5 \cite{raffel2020exploring} established a text-to-text framework, unifying all NLP tasks as a sequence generation problem. Moreover, models like CLIP \cite{radford2021learning} utilize image-text pairs for multimodal tasks with the images captioned by natural language. In the realm of reinforcement learning, \citet{di2023towards} improves the agent by employing natural language to describe environmental states.
P5 \cite{geng2022recommendation,hua2023index,xu2023openp5} and its variants \cite{geng2023vip,hua2023up5,ji2023genrec}, further contributes to this trend by reformulating all personalized recommendation tasks as language modeling tasks via prompts. The aforementioned works collectively demonstrate that employing natural language for multimodal data representation has emerged as a prominent and promising trend.

 \raggedbottom
However, in graph machine learning, such an exploration still remains limited. Existing methods that utilize LLMs for graph can be roughly categorized into two types: 1) Combining LLMs and GNNs, where the LLM acts as a feature extractor or data augmentation module to enhance the downstream GNNs \cite{he2023explanations,mavromatis2023train,zhao2022learning}. These methods often require training multiple models, incurring significant computational overhead and tend to easily inherit drawbacks of GNNs such as over-smoothing \cite{cai2020note}. 2) Purely relying on Transformers but necessitating novel designs of token embedding for nodes and edges \cite{kim2022pure} or creating complex graph attention modules to learn structural information \cite{dwivedi2020generalization,nguyen2022universal}. This type of method demands local attention calculation on every node during each optimization step, leading to considerable computation costs and thus limiting each node's scope to only 1-hop neighbors. Additionally, the complex pipeline with special attention mechanisms or token representations prevents the model from directly observing and learning structural information like GNNs, thus restricting further improvement on performance.

To address the issues of LLM-based graph learning and bridge the gap between languages and graphs, we propose \textbf{InstructGLM} (\textbf{Instruct}ion-finetuned \textbf{G}raph \textbf{L}anguage \textbf{M}odel). Given that LLMs have succeeded in many AI domains, we aim to answer the question: Besides CNNs and RNNs, can LLMs also replace GNNs as the foundation model for graph machine learning? Intuitively, as one of the most expressive medium, natural language is adept at describing complex structures such that InstructGLM owns the following advantages over GNNs:

\begin{enumerate}[label=\arabic*)]
 \vspace{-4pt}
 \item \textit{Flexibility}. A natural language sentence is capable of effectively describing the connectivity at any desired hop level and intermediate paths without iterative message passing and aggregation. Even multimodal features of the nodes and edges can be directly integrated into natural language prompts, making natural language a very flexible medium to convey both structure and content on the graph.
 \vspace{-2pt}
 \item \textit{Scalability}. Injecting graph structure into multiple natural language sentences enables mini-batch training and independent gradient propagation, which facilitates scalable distributed training and low machine communication overhead for massive graphs.
 \vspace{-4pt}
 \item \textit{Compatibility}. With structure descriptions, InstructGLM is able to consistently reformulate various graph learning pipelines as language modeling tasks. This aligns well with the LLM-based multimodal processing framework, enabling the integration of graph learning with other AI domains, including vision, language, and recommendation, to build unified AI systems.
\end{enumerate}

In this paper, we focus on node classification and link prediction---two of the most fundamental tasks for graph learning. Besides, self-supervised link prediction can augment and enhance the node classification performance. 
We design a series of graph prompts for generative LLMs.
Specifically, we systematically employ natural language to describe the graphs' topological structures according to our prompts, making the graph structure clearly and intuitively provided to LLM without complex pipelines tailored to graphs. Therefore, we can handle graph tasks efficiently and succinctly by the vanilla Transformer architecture \cite{vaswani2017attention} and language modeling objective \cite{zhang2018generalized} in a generative manner. 
Overall, our contributions can be summarized as:

\begin{itemize}[leftmargin=*, topsep=2pt]
	\item Structural information is the most fundamental information for graphs, and our research shows that this fundamental information can be effectively described by languages. To the best of our knowledge, we are the first to propose purely using natural language for graph structure representation and conduct instruction tuning on generative LLMs to solve graph problems. We eliminate the requirement of designing specific complex attention mechanisms tailored for graphs. Instead, we offer a concise and efficient natural language processing interface for graph learning, which exhibits high scalability to a unified multimodal and multitask framework, aligning with the current trend across other AI domains.
	\item Inspired by various message passing mechanisms in GNNs, we have designed a series of rule-based, highly scalable instruction prompts for general graph structure representation and graph ML. Although in this paper, our focus lies in exploring instruction tuning on Large Language Models, these prompts can also be utilized for zero-shot experiments on LLMs.  
	\item We conduct self-supervised link prediction as an generic auxiliary task and further investigate its influence on the primary node classification task under a multitask instruction tuning framework. This investigation offers valuable insights into future LLM-based multitask graph learning, highlighting the importance of self-supervised link prediction in enhancing large language models' understanding of graph structures.
	\item We implement extensive experiments on three widely used graphs: ogbn-arxiv, Cora, PubMed. The results demonstrate our InstructGLM outperforms previous competitive GNN baselines and Transformers-based methods across all three datasets, achieving the top-ranked performance. LLM envisions a technical paradigm where ``everything is tokenized''. Benefiting from LLM's powerful expressive capability in representing raw data of various modality into text or non-text tokens, all types of node or edge features can essentially be transformed into LLM-compatible tokens, thereby reshaping both the graph structure and the graph attribute information into language tokens, showing the general applicability of our approach. Our experimental results validate the effectiveness of InstructGLM under general graph problem settings and emphasize the trend of utilizing generative LLMs as the new foundational model for graph machine learning.
\end{itemize}

\section{Related Work}
\subsection{GNN-based Methods}
Graph Neural Networks (GNNs) \cite{zhou2020graph,wu2020comprehensive,han2023alternately,wu2022gtnet} have been dominant in graph machine learning for a long period. Leveraging message passing and aggregation, GNNs excel in simultaneously learning node features and graph topology. Overall, GNNs with various message passing mechanisms can be categorized as spatial-based ones \cite{hamilton2017inductive,velivckovic2017graph,xu2018powerful,monti2017geometric} and spectral-based ones \cite{kipf2016semi,defferrard2016convolutional, yadati2019hypergcn}. Inherently, GNNs easily suffer from over-smoothing \cite{cai2020note}, with various regularization techniques such as MixHop, Jump Knowledge and EdgeDrop \cite{xu2018representation,abu2019mixhop,rong2019dropedge} proposed to mitigate such an overfitting. 
Another major drawback of GNNs is their inability to directly process non-numeric raw data such as text or images, requiring additional feature engineering techniques like BoW, TF-IDF, or Skip-gram as a preprocessing step \cite{wang2021bag}. Its lack of compatibility with existing large-scale generative models presents a significant challenge for integration with other AI domains such as vision and language into a unified intelligent system.

\subsection{Transformers-based Methods}
Attention-based Transformer models can be utilized for graph processing by representing nodes and edges as distinct tokens \cite{muller2023attending}. However, it is computationally intensive for handling large-scale graphs and the global attention mechanism can not effectively capture the graph's topology \cite{kim2022pure}.
To mitigate the issue, some methods incorporate graph structure information into attention matrices \cite{ying2021transformers,park2022grpe}, while others restrict attention to local subgraphs \cite{nguyen2022universal} or ingeniously design graph orthogonal vectors for node and edge tokens \cite{kim2022pure}. These newly designed complex pipelines result in indirect representation of graph structure and significantly increase the learning difficulty.
\citet{zhang2021graphprompt} utilizes natural language templates for biological concept linking \cite{sokal1970biological,wang2023exploring}. However, it can be difficult to be extended beyond classification due to the use of encoder-only model \cite{liu2019roberta}. Additionally, its natural language templates are not designed for general graph learning thus not as expressive and flexible to serve as a foundation model for graph learning.

\subsection{Fuse GNN and Transformers}
GNNs excel at learning structure, while Transformers are proficient in capturing multi-modality features. To combine the advantages of both, \citet{chien2021node} and \citet{duan2023simteg} utilizes multi-scale neighborhood prediction and LoRA \cite{hu2021lora}, respectively, to incorporate language models for generating structure enhanced feature for downstream GNNs. \citet{mavromatis2023train} employs GNNs to perform knowledge distillation on LMs, \citet{zhao2022learning} trains GNNs and LMs iteratively in a variational inference framework, while \citet{rong2020self} attempts to replace attention heads with GNNs to better capture global information.
The main drawback of the aforementioned methods is the lack of decoupling between Transformers and GNNs, results in training multiple models and incurs significant computational overhead \cite{nguyen2022universal}. Moreover, the model performance is still susceptible to inherent issues of GNNs, such as over-smoothing \cite{yang2020revisiting} and the pipeline of multi-model training is usually very complex compared to
the simplicity of a single generative LLM framework.

\subsection{Large Language Model based Methods}
Inspired by the remarkable zero-shot capabilities, leveraging LLMs in graph problems has attracted considerable attention. Existing works have included utilizing LLM to select the most suitable graph processor based on the query \cite{zhang2023graph}, employing LLM's zero-shot explanations for data augmentation to obtain advanced graph features \cite{he2023explanations}, generating prompts and benchmarks for graph construction, evaluation, biology and structural reasoning \cite{han2023pive,jiang2023structgpt,qian2023can,guo2023gpt4graph}. There are three works sharing similarities with ours. 
\citet{guo2023gpt4graph} attempts to complete graph tasks by describing graphs. However, it uses complex formal languages like \cite{brandes2013graph,himsolt1997gml} but not flexible natural language. \citet{wang2023can} and \citet{chen2023exploring} both explore using natural language with LLM for graph problems, with \cite{wang2023can} focusing on mathematical problems on small graphs while \cite{chen2023exploring} concentrating on node classification in Text-Attributed Graphs (TAGs) \cite{hu2020open}. In comparison, our natural language instruction prompts exhibit better scalability, applicable to both small and large graphs and not limited to specific graph type. Besides, the three related works only explored the basic capability of LLM for graph tasks in a zero-shot setting. Their performance does not surpass GNN baselines for the most of time with the model freezed, merely demonstrating the potential of LLM as an optional candidate for graph tasks. By contrast, we successfully bridge this gap by conducting instruction tuning on generative LLMs with simple prompts, achieving experimental results that surpass all competitive GNN baselines.

\section{InstructGLM}
In this section, we introduce \textbf{InstructGLM}, a framework utilizing natural language to describe both graph structure and meta features of node and edge for generative LLMs and further addressing graph-related tasks by instruction-tuning. We start with notation setup, followed by outlining the principles behind the design of instruction prompts, and then present a detailed illustration of the pipeline.

\begin{figure*}
\vspace{-40pt}
	\centering
	\includegraphics[width=0.95\linewidth]{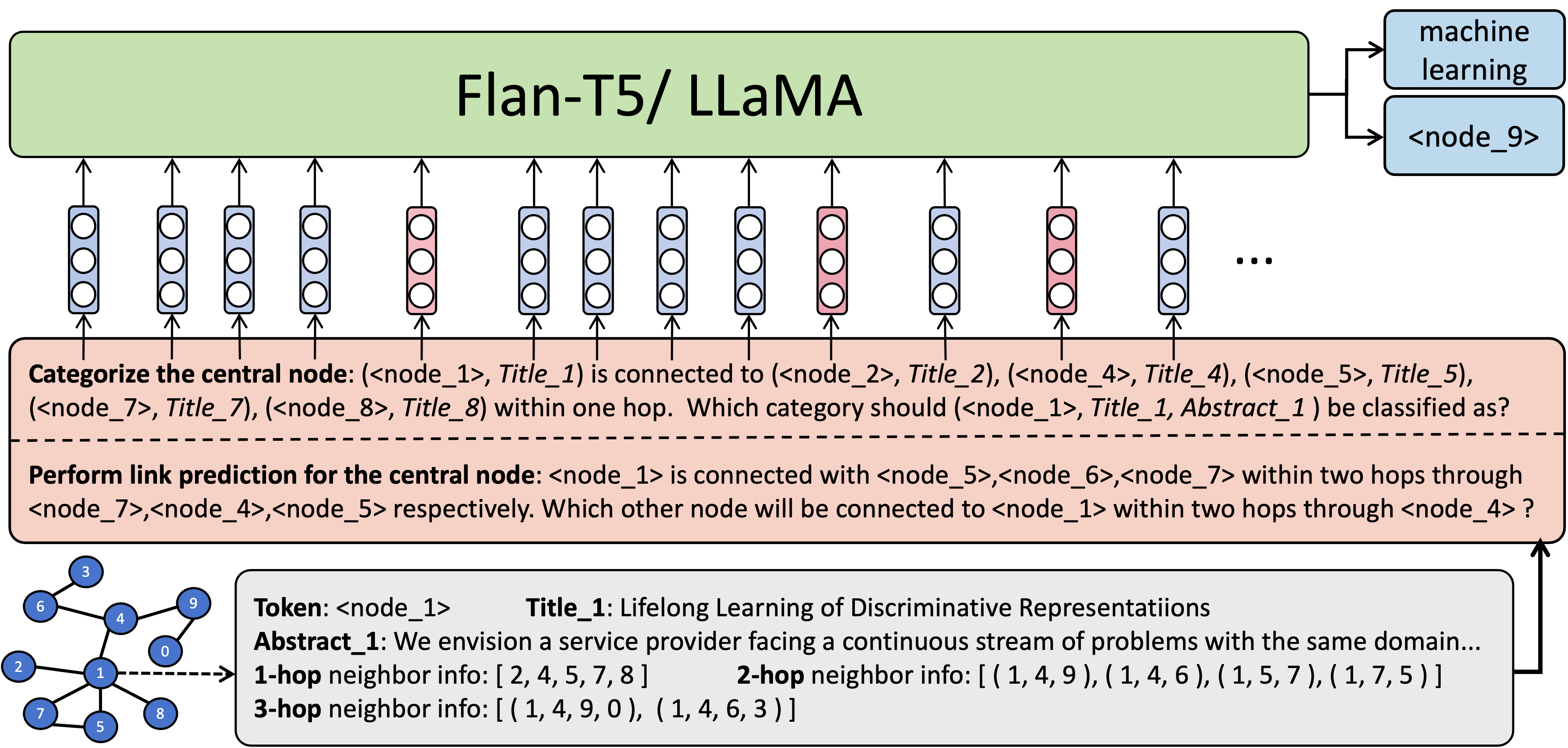}
	\caption{Illustration of InstructGLM. We use graph prompts to describe each node's multi-hop connectivity and meta features in a scalable mini-batch manner, conveying graph structure concisely and intuitively by pure natural language for learning. Subsequently, we instruct LLMs to generate responses for various graph tasks in a unified language modeling pipeline. We also expand the LLM's vocabulary by creating a new and unique token for each node. More specifically, we set the graph's inherent node feature vectors (e.g. BoW, OGB) as the embedding for these new tokens (depicted as red vectors in the figure) and employ the LLM's pre-trained embedding (depicted as blue vectors in the figure) for natural language tokens.}
	\label{fig_2}
 \vspace{-15pt}
\end{figure*}

\subsection{Preliminary}
Formally, a general graph can be represented as $\mathcal{G}=(\mathcal{V},\mathcal{A},E,\{\mathcal{N}_{v}\}_{v\in\mathcal{V}},\{\mathcal{E}_{e}\}_{e\in E} )$, where $\mathcal{V}$ is the set of nodes, $E \subseteq \mathcal{V} \times \mathcal{V}$ is the edge set, $\mathcal{A} \in \{0,1\}^{|\mathcal{V}| \times |\mathcal{V}|}$ is the adjacent matrix, $\mathcal{N}_{v}$ is the node feature of $v \in \mathcal{V}$ and $\mathcal{E}_{e}$ is the edge feature of $e\in E$. It is worth noting that the node features and edge features can be in various modalities and in diverse forms. For example, node features can be textual information in citation networks, visual images in photography graphs, user profiles in social networks, and even video or audio signals in movie networks. Similarly, edge features can be user friendships in social networks, or product reviews in user-item interaction graph of recommender systems, etc.

\subsection{Instruction Prompt Design}
In order to comprehensively convey the structure information of a graph and ensure the adaptability of the created instruction prompts to various types of graphs, we have systematically designed a set of graph description prompts centered around a central node. We mainly consider the following three questions when designing the prompts: 
\textbf{\romannumeral1)} What is the largest hop level of neighbor information about the central node in the prompt?
\textbf{\romannumeral2)} Does the prompt include meta node features or edge features?
\textbf{\romannumeral3)} For prompts with large ($\ge2$) hop level neighbors about the central node, does the prompt encompass information about the intermediate nodes or paths along the corresponding connecting route?

Regarding question \textbf{\romannumeral1)}, prompts can be classified into two types: those exclusively contain 1-hop connection information, and those with a maximum of 2-hop or 3-hop connection details. 
Prior works have shown that utilizing up to 3-hop connectivity is sufficient for excellent performance \cite{hamilton2017inductive,velivckovic2017graph,kipf2016semi}, while information beyond 3-hop typically owns a minor impact on improvement and might even lead to negative effects \cite{zhang2021evaluating,cai2020note}.  Therefore, the maximum level of neighbor information included in the prompts is up to three. However, benefiting from the flexibility of natural language, our designed prompts can actually accommodate structural information of any hop level. Regarding question \textbf{\romannumeral2)} and \textbf{\romannumeral3)}, there are two possible scenarios for each question, i.e., if or not to include the node or edge meta features in the prompt, and if or not to include the intermediate connecting paths in the prompt. 

We then denote an instruction prompt as $\mathcal{T}(\cdot)$ such that $\mathcal{I}=\mathcal{T}(v,\mathcal{A},\{\mathcal{N}_{v}\}_{v\in\mathcal{V}},\{\mathcal{E}_{e}\}_{e\in E})$ is the input natural language sentence to LLM and $v$ is the \textbf{central node} of this prompt. For instance, the simplest form of a graph description prompt containing at most 2-hop neighbor information is:
\begin{equation}
\begin{split}
\mathcal{T}(v,\mathcal{A})= &\{v\} \mbox{ is connected with } \\ &\{[v_2]_{v_2 \in \mathcal{A}^{v}_2}\} \mbox{ within two hops.} \nonumber
\end{split}
\end{equation}
\noindent
while its most detailed form which includes node features, edge features and the corresponding intermediate paths should be: 
\begin{equation}
\begin{split}
\mathcal{T}(v,&\mathcal{A},\{\mathcal{N}_{v}\}_{v\in\mathcal{V}},\{\mathcal{E}_{e}\}_{e\in E})=\{(v, \mathcal{N}_v)\} 
 \mbox{ is}\\ &\mbox{ connected with } 
\{[(v_2,\mathcal{N}_{v_2})]_{v_2 \in \mathcal{A}^{v}_2}\} \\ &  \mbox{ within two hops through } \{[(v_1,\mathcal{N}_{v_1})]_{v_1 \in \mathcal{A}^{v}_1}\} \\ &\mbox{ and featured paths }\{[(\mathcal{E}_{(v,v_1)},\mathcal{E}_{(v_1,v_2)})]\\ &_{v_1 \in \mathcal{A}^{v}_1 \,, \, v_2 \in \mathcal{A}^{v_1}_1}\}\mbox{, respectively.} \nonumber
\end{split}
\end{equation}
where $\mathcal{A}^{v}_k$ represents the list of node $v$'s $k$-hop neighbor nodes. Essentially, the above prompt should contain all 2-hop paths with node and edge features like $(v,\mathcal{N}_{v})\stackrel{\mathcal{E}_{(v,v_1)}}{\longrightarrow} (v_1,\mathcal{N}_{v_1})\stackrel{\mathcal{E}_{(v_1,v_2)}}{\longrightarrow}(v_2,\mathcal{N}_{v_2})$ centering at node $v$. All our instruction prompts are summarized in Appendix \ref{appendix:instructions}.

\subsection{Generative Instruction Tuning for Node Classification}
In prompt engineering \cite{li2021prefix,lester2021power,shin2020autoprompt} or in-context learning \cite{dong2022survey}, pretrained models are usually frozen. Instruction Tuning \cite{wei2021finetuned, chung2022scaling}, however, directly conveys the requirements of downstream tasks to pretrained models by fusing the original input data with task-specific instructional prompts under the framework of multi-prompt training. This facilitates remarkably effective fine-tuning, especially when coupled with human feedback (RLHF) \cite{ouyang2022training}. Instruction Tuning has already become an indispensable technique for fine-tuning the most powerful large language models.

In this paper, we propose InstructGLM as a multi-prompt instruction-tuning framework tailored for graph learning. Specifically, We utilize a generative large language model, either with an encoder-decoder or a decoder-only architecture, as the backbone. And then we fuse all of our designed instruction prompts, which are spanning at different hop levels with diverse structural information, together as input to the LLM, enabling mutual enhancement among the instructions. By exclusively using natural language to depict graph structures, we succinctly present the graph structure to the LLM and provide a pure NLP interface for all graph-related tasks, making them solvable via a unified pipeline in generative manner. Worth noting that we concentrate on solving node classification task in this study. We train InstructGLM to strictly generate the category label in natural language, and the prevalent Negative Log-Likelihood (i.e. NLL) Loss in language modeling are employed as our objective function.

Given $\mathcal{G}=(\mathcal{V},\mathcal{A},E,\{\mathcal{N}_{v}\}_{v\in\mathcal{V}},\{\mathcal{E}_{e}\}_{e\in E} )$ and a specific instruction prompt $\mathcal{T} \in \{\mathcal{T}(\cdot)\}$, we denote $\mathbf{x}$ and $\mathbf{y}$ as the LLM's input and target sentence, respectively. Then our pipeline can be formed as:
\begin{equation}
P_{\theta}\left(\mathbf{y}_{j} \mid \mathbf{x},\mathbf{y}_{<j}\right)=\mbox{LLM}_{\theta}\left(\mathbf{x},\mathbf{{y}_{<j}}\right), \nonumber
\end{equation}
\begin{equation}
\mathbf{x}=\mbox{Concatenate}(\mathcal{P}; \mathcal{I}; \mathcal{Q}) \nonumber
\end{equation}
\begin{equation}
\mathcal{L}_{\theta}=-\sum_{j=1}^{|\mathbf{y}|} \log P_{\theta}\left(\mathbf{y}_{j} \mid \mathbf{x},\mathbf{y}_{<j}\right) \nonumber
\end{equation}

\noindent
where $\mathcal{I}=\mathcal{T}(v,\mathcal{A},\{\mathcal{N}_{v}\}_{v\in\mathcal{V}},\{\mathcal{E}_{e}\}_{e\in E})$ is the graph structure description centering at node $v \in \mathcal{V}$, $\mathcal{L}$ denotes the NLL loss, $\mathcal{P}$ and $\mathcal{Q}$ are the task-specific instruction prefix and query. Specifically, for node classification, we design $\mathcal{P}$ and $\mathcal{Q}$ for node classification as follows: $\mathcal{P}=$ {\color{blue}`Classify the central node into one of the following categories: [<\textit{All category}>]. Pay attention to the multi-hop link relationships between the nodes.'} and $\mathcal{Q}=$ {\color{purple}`Which category should $\{v\}$ be classified as?'}. More details of the pipeline are depicted in \textbf{Figure \ref{fig_2}}.
 
Our InstructGLM actually shares essential similarities in mechanisms with various GNNs, thus inheriting their advantages. First, similar to MixHop \cite{abu2019mixhop}, which performs graph convolutions on subgraphs extracted at different hop levels, we mix prompts with diverse hop-level information 
during training. Second, Jumping Knowledge \cite{xu2018representation} combines outcomes from different convolution layers via jump connections, which is aligned with our prompts featuring intermediate information and high-hop-level neighbors.
Additionally, due to LLM's input length limit, similar to GraphSAGE \cite{hamilton2017inductive}, we conduct neighbor sampling for the central node when filling the prompts to form a mini-batch training. This operation also resembles graph regularization techniques like DropEdge \cite{rong2019dropedge} for preventing over-smoothing \cite{chen2020measuring}. 
Moreover, InstructGLM surpasses GNNs in expressiveness. Even a single graph description that contains intermediate paths and $k$-hop neighbor information is equivalent to a $k$-layer GNN in expressiveness. Therefore, InstructGLM can readily accommodate the inductive bias of graph tasks without any alterations on LLM's architecture and pipeline. For instance, since our inputs are centralized graph descriptions that directly exhibit the corresponding multi-hop neighbors, self-attention \cite{vaswani2017attention} applied on such inputs can be seen as an advanced multi-scale weighted average aggregation mechanism of GATs \cite{velivckovic2017graph,li2021training}, facilitating InstructGLM to effectively grasp different neighbors' varying importance to the central node.

\subsection{Auxiliary Self-Supervised Link Prediction}
Both SuperGAT \cite{kim2022find} and DiffPool \cite{ying2018hierarchical} introduce auxiliary link prediction task, thus successfully obtain better node representations and performance for node or graph classification, demonstrating that model's comprehension of graph structure can be significantly enhanced by such an auxiliary task. Inspired by them,
also to remove the restriction that our instruction prompts can only treat labeled training nodes as central nodes in single-task semi-supervised learning, 
we introduce self-supervised link prediction as a foundational auxiliary task for InstructGLM. Given arbitrary hop level and central node, we randomly select a neighbor or non-neighbor at this hop level as the candidate. Then we instruct our model to either discriminate whether there is a connection at this hop level between the central node and the candidate node (discriminative prompt) or directly generate the correct neighbor in a generative manner (generative prompt).

Given $\mathcal{G}=(\mathcal{V},\mathcal{A},E,\{\mathcal{N}_{v}\}_{v\in\mathcal{V}},\{\mathcal{E}_{e}\}_{e\in E} )$, the pipeline of link prediction aligns exactly with node classification. The only distinction lies in the newly designed task-specific prefix and two different query templates for it. Specifically, we design $\mathcal{P}$ and $\mathcal{Q}$ for link prediction as follows: $\mathcal{P}=$ {\color{blue}`Perform link prediction for the central node. Pay attention to the multi-hop link relationships between the nodes.'}, $\mathcal{Q}_{generative}=$ {\color{purple}`Which other node will be connected to $\{v\}$ within $\{h\}$ hop?'} and $\mathcal{Q}_{discriminative}=$ {\color{purple}`Will $\{\tilde{v}\}$ be connected to $\{v\}$ within $\{h\}$ hop?'}, where $v$ is the central node, $\Tilde{v}$ is the candidate node and $h$ is the specified hop level. We enable arbitrary node to act as central node via self-supervised link prediction and ensure a multi-task multi-prompt framework.

\section{Experiments}
\subsection{Experimental Setup}
In this paper, we primarily utilize InstructGLM for node classification, and also conduct self-supervised link prediction as an auxiliary task. Specifically, we select the following three popular citation graphs: ogbn-arxiv \cite{hu2020open}, Cora and PubMed \cite{yang2016revisiting}, in which every node represents an academic paper on a specific topic, with its title and abstract included in raw text format. We use accuracy as our metrics in all experiments and employ the default numerical node embedding of the datasets to extend the LLM's vocabulary by adding node-wise new tokens. Implementation details and elaborated dataset-specific statistics are summarized in Appendix \ref{appendix:Implement} and \ref{appendix:dataset}.

\subsection{Main Results}
Our results achieve single-model state-of-the-art performance, 
surpassing all single graph learners across all three datasets, including both representative GNN models and graph Transformer models, which 
demonstrates the promising trend for large language models to serve
as the new foundation model for graph learning. 

\subsubsection{ogbn-arxiv}
For the ogbn-arxiv, we adopt the same data split as in the OGB open benchmark \cite{hu2020open}, i.e. 54\%/18\%/28\% for train/val/test splits, respectively.

\begin{table}[htbp]
	\centering
	\begin{tabular}{p{2.3cm} p{2cm}<{\centering} p{2cm}<{\centering}}
		\toprule
		Method & OGB & GIANT\\
		\midrule
		MLP & 55.50 ± 0.23 & 73.06 ± 0.11  \\
		GAMLP  & 56.53 ± 0.16 & 73.35 ± 0.08 \\
  GraphSAGE  & 71.19 ± 0.21 & 74.35 ± 0.14  \\
  GCN     & 71.74 ± 0.29  & 73.29 ± 0.01 \\
  DeeperGCN  & 71.92 ± 0.16   & --\\
  ALT-OPT   & 72.76 ± 0.00   & --\\
  UniMP   & 73.11 ± 0.20  & -- \\
  LEGNN & 73.37 ± 0.07  & -- \\
  GAT  & 73.66 ± 0.11  & 74.15 ± 0.05 \\
  AGDN   & 73.75 ± 0.21  & 76.02 ± 0.16 \\
  RvGAT  & 74.02 ± 0.18  & 75.90 ± 0.19 \\
  DRGAT    & 74.16 ± 0.07   & \underline{76.11 ± 0.09}\\
  \midrule
  CoarFormer     & 71.66 ± 0.24  & -- \\
  SGFormer   & 72.63 ± 0.13  & -- \\
  Graphormer   & 72.81 ± 0.23   & --\\
     E2EG   & 73.62 ± 0.14   & --\\
  \midrule
  \textbf{Flan-T5-base} & 73.51 ± 0.16  & 74.45 ± 0.11\\
  \textbf{Flan-T5-large} & \underline{74.67 ± 0.08}   & 74.80 ± 0.18\\
 \textbf{Llama-7b} & \textbf{75.70 ± 0.12}  & \textbf{76.42 ± 0.09} \\
		\bottomrule
	\end{tabular}
 \caption{Results on ogbn-arxiv. We report accuracy on GNNs (Top), Graph Transformers (Middle) and our InstructGLM with different backbones (Bottom).}
 \label{arxiv}
\end{table}

\newpage
We select top-ranked GNNs from the OGB Leaderboard\footnote{\href{https://ogb.stanford.edu/docs/leader_nodeprop/}{stanford-ogbn-arxiv leaderboard}}, including DRGAT, RevGAT, etc., as the baselines \cite{zhang2022graph,hamilton2017inductive,kipf2016semi,li2020deepergcn,han2023alternately,shi2020masked,yu2022label,velivckovic2017graph,sun2020adaptive,li2021training,zhang2023drgcn}. Several most powerful Transformer-based single-model graph learners like Graphormer are also considered for comparison \cite{kuang2021coarformer,wu2023simplifying,ying2021transformers,dinh2022e2eg}.

We instruction-finetune Flan-T5 \cite{chung2022scaling} and Llama-v1 (LoRA) \cite{touvron2023llama,hu2021lora} as the backbone for our InstructGLM. The experimental results in Table \ref{arxiv} demonstrate that both models outperform all the GNNs and Transformer-based methods.
Particularly, when using Llama-v1-7b as the backbone on the default OGB feature, our InstructGLM attains a \textbf{1.54\%} improvement over the best GNN method and a \textbf{2.08\%} improvement over the best Transformer-based method. Moreover, we also achieve new \textbf{SoTA} performance on another popular and advanced feature named GIANT \cite{chien2021node}, which is enhanced by graph structure information via multi-scale neighborhood prediction task during preprocessing.

\subsubsection{Cora \& PubMed}
In terms of the compared methods for Cora and PubMed datasets \cite{he2023explanations}, we select those 
top-ranked GNNs from the two corresponding benchmarks\footnote{\href{https://paperswithcode.com/sota/node-classification-on-cora-60-20-20-random}{Cora-60-20-20-random leaderboard}} \footnote{\href{https://paperswithcode.com/sota/node-classification-on-pubmed-60-20-20-random}{PubMed-60-20-20-random leaderboard}} with 60\%/20\%/20\% train/val/test splits, including Snowball, RevGAT, etc. \cite{abu2019mixhop,pei2020geom,wu2019simplifying,he2021bernnet,bo2021beyond,chen2020simple,luan2022revisiting}. Three most powerful Transformer-based single-model graph learners on the two benchmarks, i.e., CoarFormer, Graphormer, and GT \cite{dwivedi2020generalization}, are also considered as baseline for comparison.

We instruction-finetune Flan-T5 and Llama-v1 (LoRA) as the backbone for our InstructGLM. The experimental results in Table \ref{Cora} show that our InstructGLM outperforms all the GNNs and Transformer-based methods. Specifically, InstructGLM achieves a \textbf{1.02\%} improvement over the best GNN method and a \textbf{2.08\%} improvement over the best Transformer-based method on Cora dataset, while also achieves a \textbf{3.18\%} improvement over the best GNN and a \textbf{4.87\%} improvement over the best Transformer-based method on PubMed dataset.

\begin{table}[htbp]
	\centering
	\begin{tabular}{p{2.3cm} c c}
		\toprule
		Method & Cora& PubMed\\
		\midrule
  MixHop & 75.65 ± 1.31  & 90.04 ± 1.41 \\
  GAT & 76.70 ± 0.42 & 83.28 ± 0.12\\
  Geom-GCN & 85.27 ± 1.48 & 90.05 ± 0.14 \\
  SGC-v2 & 85.48 ± 1.48&  85.36 ± 0.52 \\
  GraphSAGE & 86.58 ± 0.26  & 86.85 ± 0.11\\
  GCN & 87.78 ± 0.96   & 88.90 ± 0.32\\
  BernNet& 88.52 ± 0.95& 88.48 ± 0.41\\
  FAGCN & 88.85 ± 1.36 & 89.98 ± 0.54 \\
  GCNII& 88.93 ± 1.37 & 89.80 ± 0.30\\
  RevGAT & 89.11 ± 0.00 & 88.50 ± 0.05\\
Snowball-V3 & 89.59 ± 1.58 & 91.44 ± 0.59 \\
ACM-GCN+ & \underline{89.75 ± 1.16} & 90.96 ± 0.62\\
  \midrule
  Graphormer & 80.41 ± 0.30  &88.24 ± 1.50\\
  GT&  86.42 ± 0.82&  88.75 ± 0.16\\
  CoarFormer & 88.69 ± 0.82& 89.75 ± 0.31  \\
\midrule
   \textbf{Llama-7b} & 87.08 ± 0.32  & 93.84 ± 0.25 \\
  \textbf{Flan-T5-base} & \textbf{90.77 ± 0.52}  & \underline{94.45 ± 0.12} \\
  \textbf{Flan-T5-large} & 88.93 ± 1.06  & \textbf{94.62 ± 0.13} \\
		\bottomrule
	\end{tabular}
 \caption{Results on Cora and PubMed. We report accuracy on GNNs (Top), Graph Transformers (Middle) and our InstructGLM with different backbones (Bottom).}
	\label{Cora}
\end{table}
\vspace{-15pt} 

\subsection{Ablation Study}

In our experiments, two crucial operations contributing to the outstanding performance of InstructGLM in node classification task are \textbf{1)} multi-prompt instruction-tuning, which provides multi-hop graph structure information to the LLM, and \textbf{2)} the utilization of self-supervised link prediction as an auxiliary task. To validate the impact of the two key components on model performance, we conduct ablation experiments on all three datasets, the results are shown in Table \ref{Ablation}.

\begin{table*}[htbp]
\vspace{-36pt}
	\centering
	\begin{tabular}{ccccc}
		\toprule
		\multirow{2}{*}{Hop Info} &\multirow{2}{*}{Link Prediction} & \multicolumn{1}{c}{ogbn-arxiv} & \multicolumn{1}{c}{Cora}& \multicolumn{1}{c}{PubMed}   \\
  \cmidrule(lr){3-5}
		&& Llama-v1-7b & Flan-T5-base & Flan-T5-base \\
		\midrule
Multi-hop		 &w/  & \textbf{75.70\%}& \textbf{90.77\%} & \textbf{94.45\%}  \\
Multi-hop		 &w/o  & 75.37\% & 87.27\% &  94.35\%\\
 1-hop & w/o  & 75.25\% & 86.90\% & 94.30\%\\
 Structure-Free-Tuning & w/o  & 74.97\% & 75.65\% & 94.22\% \\
		\bottomrule
	\end{tabular}
 	\caption{Ablation Study Results. In particular, since Cora is equipped with the sparsest semantic feature (Bag of Words) among the three datasets (ogbn-arxiv with Skip-gram and PubMed with TF-IDF.), we can observe that introducing multi-hop structural information provides the greatest performance gain on Cora.}
	\label{Ablation}
 \vspace{-10pt}
\end{table*}

Regarding the \textit{Hop Info} column, \textit{Structure-Free-Tuning} indicates fine-tuning the model on titles and abstracts of the nodes, while \textit{1-hop} and \textit{Multi-hop} mean that we utilize prompts that merely include information from 1-hop neighbors and prompts that include information from neighbors with higher hop levels, respectively. The experimental results show that incorporating multi-hop information and including link prediction task can both enhance the model's performance for node classification.

\subsection{Instruction Tuning at Low Label Ratio}
In previous experiments, our data splits all ensured a relatively high ratio of labeled training nodes. To further investigate the scalability and robustness of our InstructGLM, we conduct experiments on the PubMed dataset using its another widely-used splits with extremely low label ratio. Specifically, we have only 60 training nodes available in this setting thus the label ratio is \textbf{0.3\%}.

\label{appendix:small_pub}
 \begin{table}[htbp]
	\centering
 \vspace{-5pt}
	\begin{tabular}{p{4cm} p{2cm}<{\centering}}
		\toprule
		Method & Accuracy\\
		\midrule
  GraphSAGE & 76.8 ± 0.9\\
  GAT & 79.0 ± 1.4   \\
 Snowball& 79.2 ± 0.3\\
GCN & 80.4 ± 0.4  \\
SuperGAT& 81.7 ± 0.5 \\
ALT-OPT & 82.5 ± 1.7  \\
GRAND& 82.7 ± 0.6 \\
SAIL & 83.8 ± 0.1  \\
  \midrule
  ANS-GT & 79.6 ± 1.0 \\
  NodeFormer & 79.9 ± 1.0  \\
  SGFormer& 80.3 ± 0.6   \\
  \midrule
  \textbf{Llama-7b} & 85.1 ± 0.6  \\
  \textbf{Flan-T5-base} & \underline{88.2 ± 0.3}  \\
  \textbf{Flan-T5-large}& \textbf{89.6 ± 0.4}  \\
		\bottomrule
	\end{tabular}
 \caption{Results on PubMed with 60 training nodes: accuracy on GNNs (Top), Graph Transformers (Middle) and InstructGLM with different backbones (Bottom).}
	\label{PubMed-20}
 \end{table}
\vspace{-2ex}

We consider top-ranked GNNs from the corresponding leaderboard\footnote{\href{https://paperswithcode.com/sota/node-classification-on-pubmed-with-public}{PubMed-Planetoid leaderboard}}, including SAIL, ALT-OPT, GRAND, etc., as the GNN baselines \cite{luan2019break,kim2022find,feng2020graph,han2023alternately,yu2022sail}. We also include the three most outstanding Transformer-based graph learners under this dataset setting, i.e., ANS-GT, NodeFormer and SGFormer \cite{zhang2022hierarchical,wu2022nodeformer,wu2023simplifying}.
We then instruction-finetune Flan-T5 and Llama as the backbone for our InstructGLM. Experimental results in Table \ref{PubMed-20} show that InstructGLM outperforms all GNNs with an improvement of \textbf{5.8\%} against the best GNN baseline. It also surpasses the best Transformer-based model by \textbf{9.3\%} and achieves new \textbf{SoTA} performance on the leaderboard, demonstrating the data-efficiency of InstructGLM.

\section{Conclusions and Future Work}
To the best of our knowledge, this work is the first attempt to represent graph structure via natural language description and then further perform instruction-tuning on generative LLMs for graph learning tasks, demonstrating the huge potential of LLMs as the new foundation model for graph ML. 
Our InstructGLM outperforms all single-model GNNs and Graph Transformers on ogbn-arxiv, Cora and PubMed datasets. Moreover, benefiting from our highly scalable instruction prompts and unified generative pipeline applicable to multi-modality data, InstructGLM can be readily extended to valuable future works along four directions: \textbf{1)} Leveraging LLMs to generate improved features like TAPE, SimTeG \cite{he2023explanations,duan2023simteg} and instruction prompts \cite{wei2022chain} for InstructGLM; \textbf{2)} Enhancing InstructGLM with knowledge distillation \cite{mavromatis2023train} and iterative training \cite{zhao2022learning} frameworks; \textbf{3)} Deploying InstructGLM on more graph tasks such as question answering on knowledge graphs \cite{chen2023knowledge}; \textbf{4)} Extending InstructGLM to other languages beyond natural language under the premise that ``everything is tokenized,'' to include visual tokens, acoustic tokens, other multi-modality tokens, or even domain specific languages or tokens \cite{li2024formalllm} such as chemical languages. Detailed future works are summarized in Appendix Section \ref{appendix:Future}. Overall, our InstructGLM provides a powerful NLP interface for graph machine learning, with generative LLMs and natural language as the driving force, it further contributes to the trend of unifying foundational model architecture and pipeline across multiple AI domains for the AGI pursuit.

\section*{Limitations}
The primary limitation of our InstructGLM lies in the input token limit of the large language model (LLM). For example, Flan-T5 can only accept a maximum sentence input length of 512, while Llama allows for 2048. When dealing with large-scale graphs, the instruction prompts we construct may not encompass all high-order neighbors within a single natural language sentence due to the limitations of sentence length. The simplest solution to this problem is to construct multiple graph description sentences for each training node (central node) to enumerate all possible neighbors at corresponding hop level. However, this leads to a rapid increase in the training data volume. 
In this work, learning from GraphSAGE \cite{hamilton2017inductive}, we repeatedly perform random sampling from the multi-hop neighbor lists of the central node until the sentence length reaches the input token limit to mitigate this issue. Despite our implementation achieving impressive results, we believe that improved neighbor sampling and selection strategies can help InstructGLM better address graph-related tasks, especially in the context of applications involving extremely large-scale graphs like knowledge graphs \cite{pan2023unifying}.

\section*{Ethics Statement}
Our method is proposed to provide a powerful natural language processing interface for graph machine learning tasks. Under normal and appropriate usage circumstances, there is no obvious evidence or tendency that our method will lead to significant negative societal impacts.

\bibliography{custom}

\newpage
\appendix
\section*{APPENDIX}
\section{Implementation Details}
\label{appendix:Implement}
We employ a multi-prompt instruction-tuning framework for all of our experiments and report test accuracy as our metric. Also, we employ a simple MLP over the default feature embedding of the node tokens to align their dimension with the natural language word token embeddings. All of all our experiments are conducted on four 40G A100 GPUs.


For ogbn-arxiv dataset, we adopt the same dataset splits as in the OGB open benchmark \cite{hu2020open}, which is 54\%/18\%/28\%. It takes 3.5 hours per epoch for Flan-T5-Large and 6 hours per epoch for Llama-7b during training.
For Cora and PubMed datasets, we use the version that contains raw text information proposed in \cite{he2023explanations} and employ a 60\%/20\%/20\% train/val/test splits for our experiments. It takes about 1.5 hours per epoch for Flan-T5-Large (770M) and 2.5 hours per epoch for Llama-v1-7b-LoRA (18M) during training. 

To investigate InstructGLM's performance under low-label-ratio training setting, following \citet{yang2016revisiting}, we conduct further experiments on the PubMed dataset with the fixed 20 labeled training nodes per class at a 0.3\% label ratio, and it takes about 5 minutes per epoch for Flan-T5-Large and 15 minutes per epoch for Llama-v1-7b during training due to limited labeled data. 

For both normal setting and low-label-ratio setting, the inference time is about 35ms on Flan-T5-Large and 450ms on Llama-7b per graph prompt sentence. 

In terms of hyper-parameter selection, we perform grid search within the specified range for the following parameters: (learning rate: {1e-5, 3e-5, 8e-5, 1e-4, 3e-4, 1e-3}), (batch size: {32, 64, 128, 256, 512}). We employed the AdamW \cite{loshchilov2017decoupled} optimizer with a weight decay at 0. All experiments are conducted with 4 epochs. 

\section{Dataset Statistics}
\label{appendix:dataset}
The detailed statistics of the datasets are shown in Table 5.

\section{Detailed Discussions on Future Work}
\label{appendix:Future}
Potential valuable future work can be explored along three dimensions:

\begin{itemize}[leftmargin=*, topsep=0pt]
	\item For TAGs, our experiments only used the default OGB-feature embeddings. Future work can consider using more advanced TAG-related embedding features such as LLM-based features like TAPE \cite{he2023explanations} and SimTeG \cite{duan2023simteg}. Additionally, leveraging LLM for Chain-of-Thought \cite{wei2022chain}, structure information summary, and other data augmentation techniques to generate more powerful instruction prompts will be a promising research direction for graph language models.
	\item InstructGLM can be integrated into frameworks like GAN and GLEM \cite{goodfellow2014generative,zhao2022learning} for multi-model iterative training, or utilize off-the-shelf GNNs for knowledge distillation \cite{mavromatis2023train}. Also, classic graph machine learning techniques like label reuse, Self-Knowledge Distillation (Self-KD), Correct \& Smooth can further enhance the model's performance.
	\item Benefiting from the high  flexibility and expressiveness of language 
 and the highly scalable design of our instruction prompts, InstructGLM can be easily extended 
 to various kinds of graphs and modalities within a unified generative language modeling framework, since ``everything can be tokenized,'' including texts, images, videos, audios and other modalities, and inserted into language prompts.
 Besides, our designed instruction prompts can be further used for inductive node classification tasks. Furthermore, with only slight modifications to the prompts, tasks such as graph classification, intermediate node or path prediction, and even relation-based question answering tasks in knowledge graphs with rich edge features can be effectively deployed.
\end{itemize}

\section{Instruction Prompts}
\label{appendix:instructions}
We present all of our designed instruction prompts. It is worth noting that we follow the following conventions when numbering the prompts:
\begin{itemize}[leftmargin=*, topsep=10pt]
	\item The length of each prompt number is 4.
	\item The first digit represents the task index, where 1 represents the node classification task and 2 represents the link prediction task.
	\item The second digit represents whether node features or edge features (such as text information) other than numerical feature embedding are used in the prompt. 1 means not used and 2 means used.
\end{itemize}



\section*{Supplementary material}

| Dataset | #Node | #Edge | #Class | Default Feature | #Features |
| --- | --- | --- | --- | --- | --- |
| ogbn-arxiv | 169,343 | 1,166,243 | 40 | Skip-gram / GIANT | 128 / 768 |
| Cora | 2,708 | 5,429 | 7 | Bag of Words | 1433 |
| PubMed | 19,717 | 44,338 | 3 | TF-IDF | 500 |

Table 5: Dataset Statistics

- The third digit represents the maximum hop order corresponding to the structural information considered in this prompt. 1 represents only the 1-hop neighbors are included, while 2 and 3 represent the structural information including 2-hop and 3-hop neighbors, respectively.

- The fourth digit represents whether the intermediate node information (i.e. the path) in the high-order connection is considered in this prompt. If the digit is even, it means that the intermediate node is considered, while an odd digit indicates otherwise.

- Specially, in node classification task, we designed a graph-structure-free prompt and numbered it as 1-0-0-0.

### D.1 Node Classification

**Task-specific prefix:**

Classify the paper according to its topic into one of the following categories:{{All Category List}}.\n Node represents academic paper with a specific topic, link represents a citation between the two papers. Pay attention to the multi-hop link relationship between the nodes.

**Prompt ID: 1-1-1-1**

Input template:

{{central node}} is connected with {{1-hop neighbor list}} within one hop. Which category should {{central node}} be classified as?

Target template: {{category}}

**Prompt ID: 1-1-2-1**

Input template:

{{central node}} is connected with {{2-hop neighbor list}} within two hops. Which category should {{central node}} be classified as?

Target template: {{category}}

**Prompt ID: 1-1-2-2**

Input template:

{{central node}} is connected with {{2-hop neighbor list}} within two hops through {{the corresponding 1-hop intermediate node list}}, respectively. Which category should {{central node}} be classified as?

Target template: {{category}}

**Prompt ID: 1-1-3-1**

Input template:

{{central node}} is connected with {{3-hop neighbor list}} within three hops. Which category should {{central node}} be classified as?

Target template: {{category}}

**Prompt ID: 1-1-3-2**

Input template:

{{central node}} is connected with {{3-hop neighbor list}} within three hops through {{the corresponding 2-hop intermediate path list}}, respectively. Which category should {{central node}} be classified as?

Target template: {{category}}

**Prompt ID: 1-2-1-1**

Input template:

({{central node}},{{text feature}}) is connected with {{1-hop neighbor list attached with text feature}} within one hop. Which category should ({{central node}},{{text feature}}) be classified as?

Target template: {{category}}

**Prompt ID: 1-2-2-1**

Input template:

({{central node}},{{text feature}}) is connected with {{2-hop neighbor list attached with text feature}} within two hops. Which category should ({{central node}},{{text feature}}) be classified as?

Target template: {{category}}

**Prompt ID: 1-2-2-2**

Input template:

({{central node}},{{text feature}}) is connected with {{2-hop neighbor list attached with text feature}} within two hops through {{the corresponding 1-hop intermediate node list attached with text feature}}, respectively. Which category should ({{central node}},{{text feature}}) be classified as?

Target template: {{category}}

**Prompt ID: 1-2-3-1**

Input template:

({{central node}},{{text feature}}) is connected with {{3-hop neighbor list attached with text feature}} within three hops. Which category should ({{central node}},{{text feature}}) be classified as?

Target template: {{category}}

**Prompt ID: 1-2-3-2**

Input template:

({{central node}},{{text feature}}) is connected with {{3-hop neighbor list attached with text feature}} within three hops through {{the corresponding 2-hop intermediate path list attached with text feature}}, respectively. Which category should ({{central node}},{{text feature}}) be classified as?

Target template: {{category}}

**Prompt ID: 1-0-0-0**

Input template:

{{central node}} is featured with its {{text feature}}. Which category should {{central node}} be classified as?

Target template: {{category}}

### D.2 Link Prediction

**Task-specific prefix:**

Perform Link Prediction for the central node:\n Node represents academic paper with a specific topic, link represents a citation between the two papers. Pay attention to the multi-hop link relationship between the nodes.

**Prompt ID: 2-1-1-1**

Input template:

{{central node}} is connected with {{1-hop neighbor list}} within one hop. Will {{candidate node}} be connected with {{central node}} within one hop?

Target template: {{yes/no}}

**Prompt ID: 2-1-1-2**

Input template:

{{central node}} is connected with {{1-hop neighbor list}} within one hop. Which other node will be connected to {{central node}} within one hop?

Target template: {{node_id}}

**Prompt ID: 2-1-2-1**

Input template:

{{central node}} is connected with {{2-hop neighbor list}} within two hops. Will {{candidate node}} be connected to {{central node}} within two hops?

Target template: {{yes/no}}

**Prompt ID: 2-1-2-2**

Input template:

{{central node}} is connected with {{2-hop neighbor list}} within two hops through {{the corresponding 1-hop intermediate node list}}, respectively. Will {{candidate node}} be connected to {{central node}} within two hops through {{the specified 1-hop intermediate node}}?

Target template: {{yes/no}}

**Prompt ID: 2-1-2-3**

Input template:

{{central node}} is connected with {{2-hop neighbor list}} within two hops. Which other node will be connected to {{central node}} within two hops?

Target template: {{node_id}}

**Prompt ID: 2-1-2-4**

Input template:

{{central node}} is connected with {{2-hop neighbor list}} within two hops through {{the corresponding 1-hop intermediate node list}}, respectively. Which other node will be connected to {{central node}} within two hops through {{the specified 1-hop intermediate node}}?

Target template: {{node_id}}

**Prompt ID: 2-1-3-1**

Input template:

{{central node}} is connected with {{3-hop neighbor list}} within three hops. Will {{candidate node}} be connected with {{central node}} within three hops?

Target template: {{yes/no}}

**Prompt ID: 2-1-3-2**

Input template:

{{central node}} is connected with {{3-hop neighbor list}} within three hops through {{the corresponding 2-hop intermediate path list}}, respectively. Will {{candidate node}} be connected to {{central node}} within three hops through {{the specified 2-hop intermediate path}}?

Target template: {{yes/no}}

**Prompt ID: 2-1-3-3**

Input template:

{{central node}} is connected with {{3-hop neighbor list}} within three hops. Which other node will be connected to {{central node}} within three hops?

Target template: {{node_id}}

**Prompt ID: 2-1-3-4**

Input template:

{{central node}} is connected with {{3-hop neighbor list}} within three hops through {{the corresponding 2-hop intermediate path list}}, respectively. Which other node will be connected to {{central node}} within three hops through {{the specified 2-hop intermediate path}}?

Target template: {{node_id}}

**Prompt ID: 2-2-1-1**

Input template:

({{central node}},{{text feature}}) is connected with {{1-hop neighbor list attached with text feature}} within one hop. Will ({{candidate node}},{{candidate text feature}}) be connected to ({{central node}},{{text feature}}) within one hop?

Target template: {{yes/no}}

**Prompt ID: 2-2-1-2**

Input template:

({{central node}},{{text feature}}) is connected with {{1-hop neighbor list attached with text feature}} within one hop. Which other node will be connected to ({{central node}},{{text feature}}) within one hop?

Target template: {{node_id}}

**Prompt ID: 2-2-2-1**

Input template:

({{central node}},{{text feature}}) is connected with {{2-hop neighbor list attached with text feature}} within two hops. Will ({{candidate node}},{{candidate text feature}}) be connected to ({{central node}},{{text feature}}) within two hops?

Target template: {{yes/no}}

**Prompt ID: 2-2-2-2**

Input template:

({{central node}},{{text feature}}) is connected with {{2-hop neighbor list attached with text feature}} within two hops through {{the corresponding 1-hop intermediate node list attached with text feature}}, respectively. Will ({{candidate node}},{{candidate text feature}}) be connected to ({{central node}},{{text feature}}) within two hops through ({{the specified 1-hop intermediate node attached with text feature}})?

Target template: {{yes/no}}

**Prompt ID: 2-2-2-3**

Input template:

({{central node}},{{text feature}}) is connected with {{2-hop neighbor list attached with text feature}} within two hops. Which other node will be connected to ({{central node}},{{text feature}}) within two hops?

Target template: {{node_id}}

**Prompt ID: 2-2-2-4**

Input template:

({{central node}},{{text feature}}) is connected with {{2-hop neighbor list attached with text feature}} within two hops through {{the corresponding 1-hop intermediate node list attached with text feature}}, respectively. Which other node will be connected to ({{central node}},{{text feature}}) within two hops through ({{the specified 1-hop intermediate node attached with text feature}})?

Target template: {{node_id}}

**Prompt ID: 2-2-3-1**

Input template:

({{central node}},{{text feature}}) is connected with {{3-hop neighbor list attached with text feature}} within three hops. Will ({{candidate node}},{{candidate text feature}}) be connected with ({{central node}},{{text feature}}) within three hops?

Target template: {{yes/no}}

**Prompt ID: 2-2-3-2**

Input template:

({{central node}},{{text feature}}) is connected with {{3-hop neighbor list attached with text feature}} within three hops through {{the corresponding 2-hop intermediate path list attached with text feature}}, respectively. Will ({{candidate node}},{{candidate text feature}}) be connected to ({{central node}},{{text feature}}) within three hops through {{the specified 2-hop intermediate path attached with text feature}}?

Target template: {{yes/no}}

**Prompt ID: 2-2-3-3**

```
Input template:

({{central node}},{{text feature}}) is connected
with {{3-hop neighbor list attached with text
feature}} within three hops. Which other node
will be connected to ({{central node}},{{text
feature}}) within three hops?

Target template: {{node_id}}
```

**Prompt ID: 2-2-3-4**

```
Input template:

({{central node}},{{text feature}}) is connected
with {{3-hop neighbor list attached with text
feature}} within three hops through {{the
corresponding 2-hop intermediate path list
attached with text feature}}, respectively.
Which other node will be connected to ({{central
node}},{{text feature}}) within three hops
through {{the specified 2-hop intermediate path
attached with text feature}}?

Target template: {{node_id}}
```



\end{document}